\documentclass[a4paper,fleqn]{cas-sc}

\usepackage[authoryear,longnamesfirst]{natbib}

\usepackage{rotating} % <-- HERE
\usepackage{multirow}
\usepackage{ulem}
\def\tsc#1{\csdef{#1}{\textsc{\lowercase{#1}}\xspace}}
\tsc{WGM}
\tsc{QE}
\tsc{EP}
\tsc{PMS}
\tsc{BEC}
\tsc{DE}
\begin{document}
\let\WriteBookmarks\relax
\def\floatpagepagefraction{1}
\def\textpagefraction{.001}
\shorttitle{Time Series Foundation Models}
\shortauthors{M. Laglil et~al.}
%\begin{frontmatter}

\title [mode = title]{Foundation Models and Fine-Tuning: Toward a New Generation of Models for Time Series Forecasting}                      
%\tnotemark[1,2]
%\tnotetext[1]{This document is the results of the research project funded by the National Science Foundation.}
%\tnotetext[2]{The second title footnote which is a longer text matter to fill through the whole text width and overflow into another line in the footnotes area of the first page.}

\author[1,2]{Morad Laglil}%[type=editor, auid=000,bioid=1, orcid=0000-0001-0000-0000]
\cormark[1]
\fnmark[1]
\ead{morad.laglil@univ-grenoble-alpes.fr}
\credit{Conceptualization of this study, Software and experimental development and testing, Writing}

%\credit{Conceptualization of this study, Methodology, Software, Writing}

\affiliation[1]{organization={Univ. Grenoble Alpes, CNRS, Grenoble INP, LIG},%, 38000 Grenoble, France},
                %addressline={Jawahar Nagar}, 
                city={Grenoble},
%               citysep={}, % Uncomment if no comma needed between city and postcode
                postcode={38000}, 
                %state={Kerala},
                country={France}}

\author[2]{Bertrand Pracca}
\credit{Conceptualization of this study}

\author[1]{Emilie Devijver}
\ead{emilie.devijver@univ-grenoble-alpes.fr}
\credit{Conceptualization of this study, Proofreading}

\affiliation[2]{organization={Savoye},
                addressline={18 bd des Gorgets}, 
                postcode={21000}, 
                city={Dijon},
                country={France}}

\author[1]{Eric Gaussier}
\ead{eric.gaussier@univ-grenoble-alpes.fr}
\credit{Conceptualization of this study, Proofreading}

\cortext[cor1]{Corresponding author}
%\fntext[fn1]{This is the first author footnote, but is common to thirdauthor as well.}
%\fntext[fn2]{Another author footnote, this is a very long footnote and it should be a really long footnote. But this footnote is not yet  sufficiently long enough to make two lines of footnote text.}

%\nonumnote{This note has no numbers. In this work we demonstrate $a_b$  the formation Y\_1 of a new type of polariton on the interface  between a cuprous oxide slab and a polystyrene micro-sphere placed on the slab.  }

%%%%%%%%%%%%%%%%%%%%%%%%%%%%%%%%%%%%%%% Abstract %%%%%%%%%%%%%%%%%%%%%%%%%%%%%%%%%%%%%%%%%%%%%%%%%%%%%%%%%%%%%%%%%%%%%%%%%
\begin{abstract}
Inspired by recent breakthroughs in large language models for natural language processing, foundation models have emerged as a promising paradigm for zero-shot time series forecasting—enabling accurate predictions on datasets never seen during pre-training. Ranging from tens to hundreds of millions of parameters, these models are pre-trained on vast and diverse collections of time series, learning generalizable representations that support both point and probabilistic forecasting. This approach alleviates the need for dataset-specific model design and manual tuning, offering a unified solution across forecasting problems.
In this work, we review the main architectures, pre-training strategies, and optimization methods underpinning these models. We further investigate post-pre-training fine-tuning of selected foundation models to enhance their performance on specific datasets. Our empirical results demonstrate that this step consistently improves forecasting accuracy over the zero-shot baseline.
\end{abstract}

%\includegraphics{figs/cas-grabs.pdf}

\begin{highlights}
\item A review of the main architectures and training strategies of time series foundation models.
\item Analysis of zero-shot vs fine-tuned performance.
\item Empirical evaluation across diverse forecasting datasets.
\end{highlights}


\maketitle

%%%%%%%%%%%%%%%%%%%%%%%%%%%%%%%%%%%%%%% Introduction %%%%%%%%%%%%%%%%%%%%%%%%%%%%%%%%%%%%%%%%%%%%%%%%%%%%%%%%%%%%%%%%%%%%%%%

\section{Introduction}
Time series forecasting has long been a cornerstone task in industry, underpinning decision-making and process optimization across domains as diverse as energy, transportation, meteorology, economics, and retail. It also remains a central focus of academic research, driving continuous progress toward more accurate predictions.

Traditional forecasting approaches rely primarily on parametric models grounded in domain expertise, such as autoregressive models (AR, \citep{box2015time}) and exponential smoothing methods \citep{article}. In contrast, modern machine learning approaches—particularly deep learning—can capture complex temporal dynamics by learning directly from data, without strong assumptions on its underlying distribution. A wide range of models, including random forests \citep{en15207547} and deep neural networks \citep{SALINAS20201181}, have thus been developed to improve forecasting accuracy and better capture the diversity of temporal patterns observed in real-world series.

More recently, \textit{foundation models} have emerged as a major breakthrough in time series forecasting. Following the definition established in \citet{DBLP:journals/corr/abs-2108-07258}, a foundation model for time series is a large-scale model pre-trained on massive and diverse temporal corpora spanning multiple domains, frequencies, and dynamics, such that it acquires general-purpose temporal representations that can be transferred to a wide range of downstream forecasting tasks (target dataset)—including previously unseen ones—without dataset-specific retraining. This distinguishes them from conventional deep learning models, which are typically trained from scratch on a single target dataset. Leveraging the growing availability of data and computational power, these models exploit inductive biases reflecting the structural properties of time series to learn such generalizable representations, enabling them to adapt to a broad range of contexts, including zero-shot forecasting.

Inspired by the remarkable successes of large language models in natural language processing, foundation models for time series can also be fine-tuned on specific datasets, allowing the knowledge acquired during pre-training to be transferred and specialized. However, this fine-tuning dimension remains largely underexplored in the literature on foundation models for forecasting. We aim to explore this dimension in this study, using as base foundation models top-performing models for time series forecasting\footnote{This paper is an extension and an update of a previous paper published at \textit{La 26ème Conférence Francophone sur l'Extraction et la Gestion des Connaissances, EGC 2026}, and available at \url{https://arxiv.org/pdf/2511.22674}.}. We focus on two main contributions:
\begin{itemize}
\item A review of the main architectures and training strategies behind several prominent foundation models for time series forecasting (Figure~\ref{timeline}).
\item An extensive empirical study of different fine-tuning approaches, including data-efficient and parameter-efficient approaches, across a diverse set of datasets spanning multiple domains, frequencies, and forecasting horizons. To the best of our knowledge, this is the first fine-tuning study of such scale in this area.
\end{itemize}

The remainder of this paper is organized as follows: 
Section~\ref{sec:bib} presents the main approaches for forecasting with time series foundation models, Section~\ref{sec:expes} reports the experimental results on fine-tuning, and Section~\ref{sec:conc} discusses remaining challenges and concludes the paper.

%%%%%%%%%%%%%%%%%%%%%%%%%%%%%%%%%%%%%%% Time Series Forecasting Problem %%%%%%%%%%%%%%%%%%%%%%%%%%%%%%%%%%%%%%%%%%%%%% 

%\section{Time Series Forecasting Problem} %\label{sec:framework}

%%%%%%%%%%%%%%%%%%%%%%%%%%%%% Foundation Models for Time Series Forecasting %%%%%%%%%%%%%%%%%%%%%%%%%%%%%%%%%%%%%%%%%%%%%% 

\section{Foundation Models for Time Series Forecasting}\label{sec:bib}
\begin{figure}[pos = t]
\centering
\includegraphics[width=16cm]{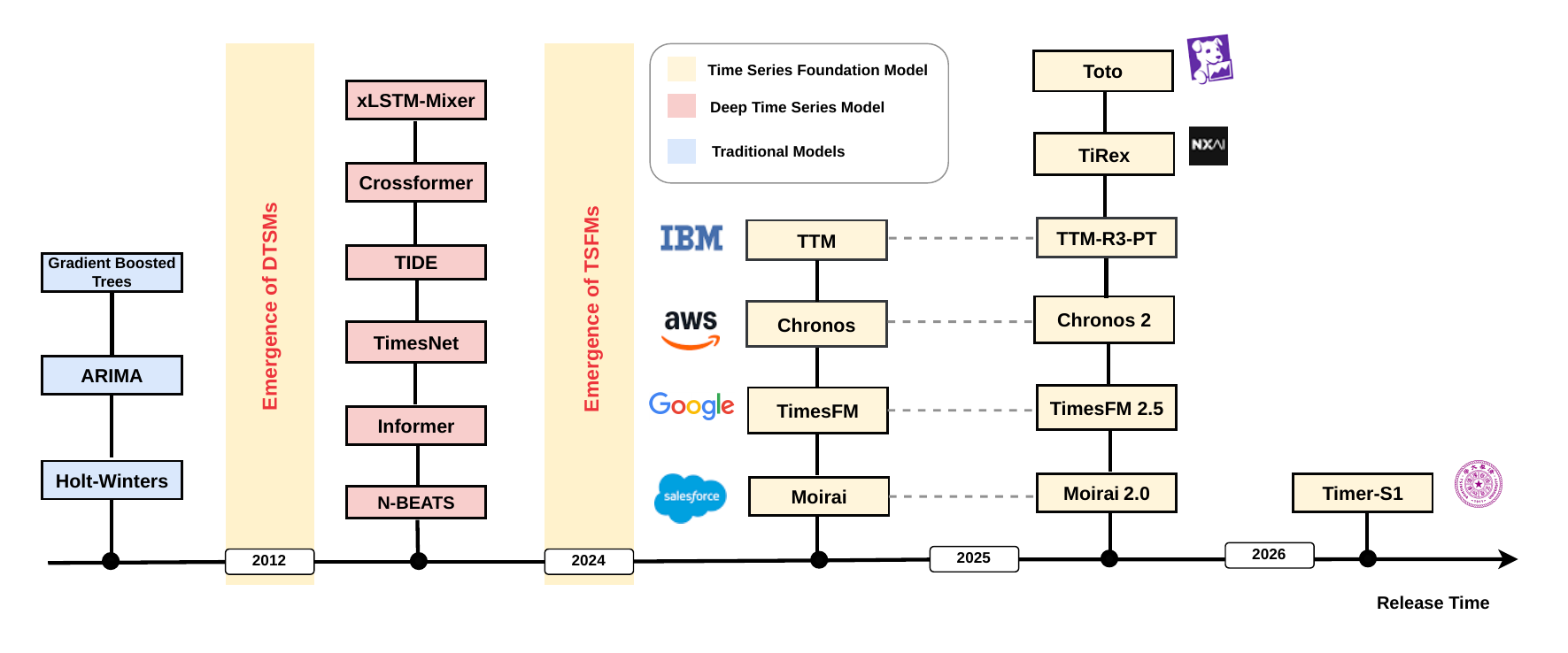} 
\caption{Timeline of representative time series forecasting models, from traditional statistical methods (ARIMA, Holt-Winters) through deep time series models (DTSMs, 2012--2024) to the emergence of pre-trained time series foundation models (TSFMs, 2024--2026). Models are positioned according to their release date, which may precede the publication of the corresponding paper or technical report.}
\label{timeline}
\end{figure}

After formally introducing the time series forecasting problem, we present in this section the main approaches for forecasting with foundation models, organized around three key components: the architecture (Section~\ref{sec:archi}), which defines the network structure and underlying computational mechanism; pre-training (Section~\ref{sec:preentr}), which involves exposing the model to large-scale datasets via self-supervised learning; and adaptation (Section~\ref{sec:ajus}), which encompasses the various fine-tuning strategies used to specialize a pre-trained model for specific tasks. Together, these three dimensions form a unified analytical framework for comparing, classifying, and deploying foundation models for forecasting. We review several representative foundation models (Figure~\ref{timeline}), presented in chronological order, that have achieved state-of-the-art performance on the GIFT-Eval benchmark \citep{aksu2024gifteval} and that illustrate a diversity of approaches across the three components outlined above.

\subsection{Time Series Forecasting Problem}\label{sec:framework}

Time series forecasting aims to predict the future values of a time-indexed series based on past observations and, possibly, exogenous variables. Formally, let $y_t \in \mathbb{R}^D$ denote the observation vector at time $t$, where $D \geq 1$ is the dimension of the series ($D = 1$ for univariate and $D > 1$ for multivariate time series), and consider a time series $\{y_t\}_{t=1}^T$. The forecasting task consists in estimating the future values $\{y_{T+1}, \ldots, y_{T+H}\}$ over a forecast horizon $H > 0$, given past observations (context) $\{y_1, \ldots, y_T\}$ and, optionally, $n_{\text{cov}}$ external covariates $\{z_1, \ldots, z_T\}$. The forecasting function is defined as a mapping:
\begin{equation*}
\begin{aligned}
f : \mathbb{R}^{T \times D} \times \mathbb{R}^{T \times n_{\text{cov}}} &\longrightarrow \mathbb{R}^{H \times D}\\
(\{y_t\}_{t=1}^T, \{z_t\}_{t=1}^T) &\longmapsto \{y_{T+1}, \ldots, y_{T+H}\}.
\end{aligned}
\end{equation*}
In the probabilistic forecasting setting, it is often more interesting to predict quantiles of the future distribution rather than single point estimates. This is usually achieved through the minimization of the quantile loss, defined, for a given quantile $q \in (0,1)$, by:
\[
\mathcal{L}_q(y, \hat{y}^q) = \big(q - \mathbf{1}_{\{y - \hat{y}^q < 0\}}\big)\, (y - \hat{y}^q),
\]
where $y$ is the true value and $\hat{y}^q$ is the predicted quantile. Equation~\ref{eq:quantile-loss} below shows how these individual losses are aggregated over multiple quantiles and across the forecast horizon $H$.

%%%%%%%%%%%%%%%%%%%%%%%%%%%%% Architecture %%%%%%%%%%%%%%%%%%%%%%%%%%%%%%%%%%%%%%%%%%%%%%%%%%%%%%%%%%%%%%%%%%%%%%ù% 
\subsection{Model Architectures}\label{sec:archi}
The foundation models considered in this work span three architectural families: Transformer-based models, recurrent models based on extended LSTMs, and MLP-Mixer-based models. In the following, we describe the design principles and key innovations of each model, relying on Figure~\ref{encodeur} to illustrate the two dominant paradigms used in time series foundation models.
%%%%%%%%%%%%%%%%%%%%%%%%%%%%% Transformer-Based Models %%%%%%%%%%%%%%%%%%%%%%%%%%%%%%%%%%%%%%%%%%%%%%%%%%%%%%%%%% 

\subsubsection{Transformer-Based Models}

\begin{figure}[pos = t]
\centering
\includegraphics[width=16cm]{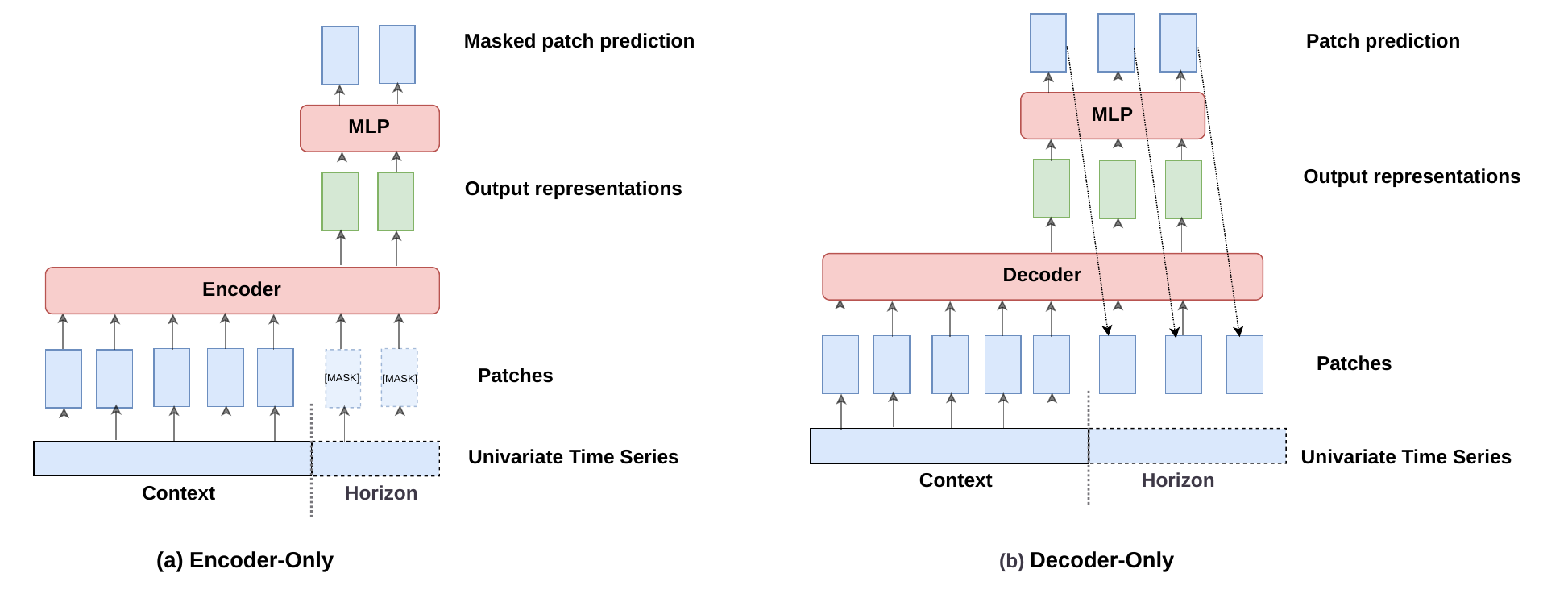} 
\caption{Illustration of the two main paradigms for time series foundation models. (a)~\textbf{Encoder-only}: future patches are masked and appended to the input; the encoder processes all patches bidirectionally and predicts the masked positions. (b)~\textbf{Decoder-only}: the model autoregressively predicts the next patch from preceding context using causal attention. Blue: input time series and its patches; red: neural layers (encoder/decoder and MLP head); green: latent representations projected by the MLP to produce the final forecasts.% Dashed arrows in~(b) indicate the autoregressive feedback of predictions into subsequent steps.
}
\label{encodeur}
\end{figure}

Most foundation models for time series forecasting rely on the \textbf{Transformer} architecture \citep{NIPS2017_3f5ee243}, originally designed for natural language processing. Their main innovation lies in the \textbf{attention mechanism}, which allows the model to assign varying importance to different parts of the input sequence. This mechanism enables the model to capture long-range dependencies between elements of the series—something earlier architectures, such as recurrent neural networks, struggled to achieve.

\cite{ansari2024chronos} introduced a model called \textbf{Chronos} (20M--710M parameters), based on an \textit{encoder-decoder} Transformer architecture (combining the two architectures shown in Figure~\ref{encodeur}). The encoder processes the past time series to extract rich representations, while the decoder generates forecasts \textit{autoregressively} from these representations, following the T5 architecture~\citep{10.5555/3455716.3455856}. The model adopts the principles of text processing but introduces an novel idea: representing time series as tokens. Specifically, each value of the series is converted into a token corresponding to the predefined interval to which it belongs. At the output, the tokens predicted by the model are mapped back to real values, for instance by assigning them the lower bound of the corresponding interval. A limitation of this approach is that the prediction range remains fixed, since the intervals are predefined, making it less suitable for time series exhibiting strong trends.

Another approach to convert a time series into units of information (or tokens) usable by Transformers consists in segmenting it into several fixed-size subsequences, called \textit{patches}. By dividing long time series into shorter patches, the model becomes better at capturing local patterns within the sequence, thereby improving its understanding of the complex dynamics it contains. Moreover, this segmentation helps reduce the computational complexity of the Transformer by decreasing the total number of tokens to process.

\citet{10.5555/3692070.3692474} applied this idea in the \textbf{TimesFM} (200M parameters) model, based on a \textit{decoder-only} Transformer architecture (Figure~\ref{encodeur}(b); in this case, the decoder is a Transformer). It relies on a causal attention mechanism, where each patch can only attend to previous patches, without ever using future information. This way, the model learns to forecast by extracting only the relevant dependencies from the past. The model adopts an \textit{autoregressive} generation scheme at inference time, producing the output patches sequentially. The size of the predicted patches is larger than that of the input patches (multi-token prediction), which enables long-horizon forecasting while reducing the number of required autoregressive iterations. TimesFM~2.5 is the latest variant, pre-trained on an extended corpus of time series.

\textbf{Moirai} (14M--311M parameters) \citep{woo2024moirai} follows the same time series segmentation principle, but introduces variable patch sizes: larger for high-frequency series (e.g., second- or minute-level) to reduce the quadratic cost of attention, and smaller for low-frequency series (e.g., daily or monthly) to better leverage the representational capacity of Transformers. To this end, Moirai learns multiple input and output projection layers associated with different patch sizes, with the appropriate size being selected manually based on the frequency characteristics of the series.

The core of the model is an encoder-only Transformer (Figure~\ref{encodeur}(a); in this case, the encoder is a Transformer) that employs a masking technique: a special \texttt{[mask]} token replaces the patches within the forecast horizon. This token has its own learnable representation, which is then decoded via an output projection to generate the corresponding forecast, thereby avoiding autoregressive generation at inference time.

Another challenge for foundation forecasting models lies in their difficulty in handling multivariate time series of arbitrary dimensions. Most existing Transformers assume independence between variables, since the number of dimensions may vary and is not known in advance. To overcome this limitation, Moirai adopts a strategy that flattens the multivariate series in order to process it as a single sequence. This strategy is paired with the introduction of variate-index encodings, allowing the model to distinguish between the different dimensions when computing attention.

\textbf{TOTO} (103M parameters) \citep{cohen2025this}, recently proposed, is a decoder-only Transformer (Figure~\ref{encodeur}(b)) that adopts an alternative approach for handling multivariate time series. Each series is first divided into multiple patches, and a two-level attention mechanism is then applied: the first captures temporal dependencies within each univariate series, while the second models interactions across series through dimensional attention \citep{zhang2023crossformer}. This architecture enables TOTO to efficiently represent both temporal and cross-variable relationships.

\textbf{Chronos~2} (120M parameters) \citep{ansari2025chronos2univariateuniversalforecasting} is an encoder-only Transformer that extends the original Chronos model with the ability to handle univariate, multivariate, and covariate-informed forecasting tasks in a zero-shot manner. The key architectural innovation is the \textit{group attention} mechanism, which enables information sharing across related time series within a batch. Each Transformer block alternates between two attention layers: a \textit{time attention} layer that aggregates information across patches within a single series, and a \textit{group attention} layer that shares information across all series belonging to the same group at each patch index (similar to TOTO). A group is a flexible notion of relatedness and can represent independent univariate series, variates of a multivariate series, or targets along with their associated covariates.

\textbf{Moirai~2.0}  (36M parameters) \citep{liu2026moirai20timeseries} is a decoder-only Transformer that simplifies and improves upon the original Moirai architecture in several key aspects. First, it replaces the masked-encoder design of Moirai with a decoder-only autoregressive architecture. Second, it adopts a single patch size instead of the multi-patch setup of its predecessor, simplifying both training and inference. For long-horizon forecasting, Moirai~2.0 employs multi-token prediction combined with an \textit{autoregressive multi-quantile decoding} strategy at inference time. Since the model outputs multiple quantiles at each step but the input expects a single value, feeding back only the median would discard uncertainty information. Instead, at each decoding step, the model expands by generating a separate forecast from each predicted quantile of the previous step, producing a larger set of candidate values. It then collapses this expanded set back into the desired quantile levels, thereby propagating the full predictive uncertainty across successive decoding steps.

\begin{figure}[pos = t]
\centering
\includegraphics[width=12cm]{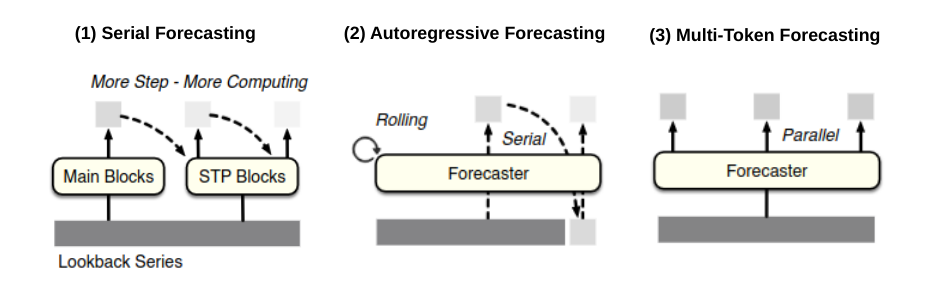}
\caption{Comparison of three forecasting strategies. (1)~\textit{Serial forecasting} (Timer-S1): the main blocks process the lookback series, then each STP block progressively predicts the next future patch with increasing computation depth. (2)~\textit{Autoregressive forecasting}: the model predicts one patch at a time and rolls its own predictions back as input. (3)~\textit{Multi-token forecasting}: Multiple patches are produced simultaneously in a single forward pass. Figure from \cite{liu2026timers1billionscaletimeseries}.}
\label{serialforecast}
\end{figure}

\textbf{Timer-S1} (8.3B parameters) \citep{liu2026timers1billionscaletimeseries} is a decoder-only Transformer that introduces \textit{Serial-Token Prediction} (STP) to address a fundamental limitation of existing forecasting strategies. The key motivation behind Timer-S1 lies in the observation that time series forecasting is an inherently \textit{serial} problem: predicting far into the future requires progressive, step-by-step reasoning, since each future value depends on all preceding estimations. In other words, the accuracy of a long-term forecast is fundamentally chained to the accuracy of shorter-term predictions — one cannot reliably predict step $k$ without first reasoning about steps $1, 2, \ldots, k{-}1$. This serial nature implies that predictions at longer horizons should undergo more computation than those at shorter horizons. Standard autoregressive models naturally respect this property by predicting one patch at a time, but they must roll their own predictions back as input, which accumulates errors over long horizons. Conversely, multi-token prediction approaches produce all future patches simultaneously but from the same representations, and they allocate the same amount of computation to every horizon, disregarding the progressive step-by-step reasoning required for forecasting (Figure~\ref{serialforecast}). Timer-S1 reconciles these two paradigms through a two-stage architecture. The first stage consists of a stack of main Transformer blocks enhanced with sparse Mixture-of-Experts (MoE) modules, where each patch token is routed to only a small subset of specialized experts, enabling the model to scale to 8.3 billion parameters while keeping inference efficient. The second stage appends a sequence of dedicated \textit{TimeSTP} blocks, each of which combines two sources of information: the embeddings produced by the preceding block and the original input patch embeddings, fused through concatenation and linear projection before being processed by an additional MoE-augmented Transformer block. This design introduces progressive serial computations for multi-horizon forecasting: each TimeSTP block predicts the next future patch, conditioned on the computation path of all previous blocks, while remaining grounded in the raw input signal. As a result, predictions at longer horizons naturally pass through more Transformer blocks than shorter-term ones, mirroring the causal dependency structure of the forecasting problem. This allows Timer-S1 to produce multiple future patches in a single forward pass without autoregressive rolling, while still respecting the sequential nature of time series forecasting. 

%%%%%%%%%%%%%%%%%%%%%%%%%%%%%%%%%%%%%% Recurrent Network-Based Models %%%%%%%%%%%%%%%%%%%%%%%%%%%%%%%%%%%%%%%%%%%

\subsubsection{Recurrent Network-Based Models}

Recurrent neural networks, particularly LSTMs \citep{10.1162/neco.1997.9.8.1735}, have been widely used for time series forecasting \citep{SALINAS20201181}. The first—and to date the only—foundation model based on a recurrent architecture is \textbf{TiRex} (35M parameters) \citep{auerTiRexZeroShotForecasting2025}. It is a decoder (Figure~\ref{encodeur}(b)) built on the sLSTM \citep{beck:24xlstm}, an extension of the classical LSTM that replaces the sigmoid activation—which constrains the output to the interval $[0, 1]$—with an exponential activation. This modification endows the network with greater expressiveness.

TiRex segments time series into patches. Unlike autoregressive decoders, which generate forecasts patch by patch (e.g., at inference time) by feeding each predicted output back as input—thereby amplifying error accumulation—TiRex treats future patches as missing values, thus preventing error propagation and improving the stability of long-horizon forecasts.

\subsubsection{MLP-Mixer-Based Models}

\textbf{TTM} (Tiny Time Mixers, 1M--36M parameters) \citep{NEURIPS2024_874a4d89} departs from the Transformer paradigm by building on the lightweight \textit{TSMixer} architecture \citep{10.1145/3580305.3599533}, which replaces the quadratic self-attention mechanism with MLP-Mixer blocks interleaved with simple gated attention modules for mixing features within patches, across patches, and across channels. This design choice drastically reduces the model size---starting from 1M parameters---while maintaining competitive forecasting accuracy. TTM segments the input series into patches and processes them through a multi-level backbone featuring \textit{adaptive patching}, where different layers operate at varying patch lengths and resolutions, inspired by hierarchical architectures from the vision domain. To handle the diversity of temporal resolutions present in the pre-training corpus, TTM introduces two additional mechanisms: \textit{diverse resolution sampling}, which augments the training data by deriving lower-resolution versions of high-frequency datasets, and \textit{resolution prefix tuning}, which prepends a learnable embedding encoding the input resolution to the patch sequence, enabling resolution-conditioned modeling. TTM-R3-PT is the latest version of this family, incorporating minor modifications to improve accuracy in high-speed forecasting.

%%%%%%%%%%%%%%%%%%%%%%%%%%%%%%%%%%%%%% Pre-training %%%%%%%%%%%%%%%%%%%%%%%%%%%%%%%%%%%%%%%%%%%

\subsection{Pre-training}\label{sec:preentr}

Pre-training constitutes an essential initial stage in building foundation models for time series. The knowledge acquired during this phase enables the models to better generalize across different contexts. However, the diversity of architectural styles and forecasting targets (e.g., encoders with masking, autoregressive decoders, multi-patch prediction) leads to a wide variety of pre-training corpora and optimization strategies in the design and development of these models.

\subsubsection{Pre-training Corpus}\label{sec:pretrain-corpus}

Pre-training corpora vary significantly across models in terms of scale, composition, and curation strategy. Chronos~\citep{ansari2024chronos} combines publicly available real-world datasets with synthetic series generated via Gaussian processes, a strategy shown to significantly improve zero-shot generalization. Its successor, Chronos~2~\citep{ansari2025chronos2univariateuniversalforecasting}, extends this corpus with datasets from the GIFT-Eval pretraining split~\citep{aksu2024gifteval} and two additional synthetic generators---TSI (combining trend, seasonality, and irregularity components) and TCM (sampling from random temporal causal graphs)---while relying entirely on synthetic data to learn multivariate and covariate-informed structures. Moirai~\citep{woo2024moirai} is trained on \textit{LOTSA}, a large multi-domain corpus in which the contribution of each dataset is capped to prevent dominant domains from overwhelming training.  Toto~\citep{cohen2025this} leverages one of the largest pre-training corpora to date, comprising over two trillion time points, of which approximately 75\% consists of proprietary observability metrics from the Datadog platform, complemented by public datasets and synthetic data generated from compositions of piecewise trends, ARMA processes, and sinusoidal patterns. Timer-S1~\citep{liu2026timers1billionscaletimeseries} is pre-trained on \textit{TimeBench}, a corpus of over one trillion time points aggregating real-world series from diverse domains (finance, IoT, meteorology, healthcare, energy) alongside collections from Chronos and LOTSA, supplemented by synthetic data. More broadly, the use of synthetically generated time series has become a recurring theme across these models, increasing the diversity of temporal patterns encountered during training and consistently improving generalization to unseen datasets.

\subsubsection{Optimization Strategies}\label{sec:pretrain-optim}

Pre-training is performed in a self-supervised manner, without requiring data annotation. The models differ primarily in their choice of loss function and supervision scheme. We organize them below according to the type of training objective, and provide a summary in Table~\ref{tab:optim_summary}.

\paragraph{Standard losses.}
Chronos~\citep{ansari2024chronos} is pre-trained by minimizing the \textit{cross-entropy loss} over tokens generated autoregressively for a fixed prediction horizon. The model learns to predict a distribution over a set of predefined tokens, which allows it to produce probabilistic forecasts (e.g., for the quantiles $0.1, 0.2, \ldots, 0.9$). Moirai~\citep{woo2024moirai} and Toto~\citep{cohen2025this} are instead pre-trained by maximizing the \textit{log-likelihood} of a parametric mixture distribution combining several families (Student's~$t$, negative binomial, log-normal, and normal) in order to model different types of temporal data.

\paragraph{Quantile loss.}
TimesFM~\citep{10.5555/3692070.3692474}, TiRex~\citep{auerTiRexZeroShotForecasting2025}, Moirai~2.0~\citep{liu2026moirai20timeseries}, Chronos~2~\citep{ansari2025chronos2univariateuniversalforecasting} and TTM-R3-PT~\citep{NEURIPS2024_874a4d89} share a common training objective: they minimize the \textit{quantile loss}, defined as:
\begin{equation}\label{eq:quantile-loss}
\mathcal{L}(y_{T+1:T+H}, \hat{y}^q_{T+1:T+H}; Q) = \frac{1}{|Q|\, H} \sum_{t=T+1}^{T + H} \sum_{q \in Q} \big(q - \mathbf{1}_{\{y_t - \hat{y}^q_t < 0\}}\big)\, (y_t - \hat{y}^q_t),
\end{equation}
where $Q$ denotes the set of quantile levels, $\hat{y}^q_t$ is the model's prediction for quantile level $q$ at time $t$, and $y_t$ is the ground-truth value. Although these models share the same loss, they differ in how they handle multi-patch forecasting. TimesFM and Moirai~2.0 use an output patch larger than the input patch, thereby reducing the number of autoregressive rolling steps required. TiRex avoids autoregressive generation entirely by treating future patches as missing data during inference, predicting all required patches in a single forward pass. To simulate this behavior during pre-training, TiRex applies a \textit{contiguous-patch masking} strategy that randomly masks consecutive patches within the input series, thereby exposing the model to inference-like scenarios with incomplete preceding context. Chronos~2 adopts an encoder-style masked prediction approach: future patches are represented as masked tokens appended to the input sequence, and the model is trained to reconstruct them, effectively framing forecasting as a fill-in-the-blank task over future horizons. Its training proceeds in two stages: a first stage with a context length of $2{,}048$ and a small number of output patches, followed by a second stage that extends the context to $8{,}192$ and increases the number of output patches, enabling the model to capture long-term seasonalities and produce long-horizon forecasts without autoregressive rolling. TTM-R3-PT directly outputs the forecast horizon without relying on autoregressive generation.

\paragraph{Composite loss.}
Timer-S1~\citep{liu2026timers1billionscaletimeseries} is trained with a composite loss summing three terms. The first two---next-token prediction (NTP) and Serial-Token Prediction (STP)---are both quantile losses applied at different forecasting horizons: NTP supervises the main backbone for one-step-ahead prediction, while STP supervises the additional TimeSTP blocks that progressively predict further into the future. The third term is an auxiliary load-balancing loss that ensures even utilization across MoE experts. A post-training stage further refines the model via continued pre-training with a horizon-decaying weighted loss to prioritize short-term accuracy, and a long-context extension from $2{,}880$ to $11{,}520$ time points.
 
\begin{table}[pos = t]
\centering
\small
\resizebox{\textwidth}{!}{
\begin{tabular}{llllll}
\hline
\textbf{Model} & \textbf{Loss Function} & \textbf{Forecasting Strategy} & \textbf{Multi-Patch Handling} & \textbf{Post-Training} & \textbf{Architecture} \\
\hline
Chronos & Cross-entropy & Autoregressive & Token-by-token generation & -- & Encoder-decoder \\
Moirai & Log-likelihood  & Masked prediction & Masked future patches & -- & Encoder-only \\
Toto & Log-likelihood  & Autoregressive & Token-by-token generation & -- & Decoder-only \\
\hline
TimesFM & Quantile loss & Autoregressive & Larger output patches & -- & Decoder-only \\
TiRex & Quantile loss & Masked prediction & Masked future patches & -- & Decoder-only (sLSTM) \\
Moirai~2.0 & Quantile loss & Autoregressive & Larger output patches & -- & Decoder-only \\
Chronos~2 & Quantile loss & Masked prediction & Masked future patches & Context extension & Encoder-only \\
TTM-R3-PT & Quantile loss & Direct forecasting & Single-pass output & -- & Encoder-decoder (MLP-Mixer) \\
\hline
Timer-S1 & Composite & Serial forecasting & Progressive STP blocks & Horizon-decaying weighted loss + context extension & Decoder-only (MoE) \\
\hline
\end{tabular}
}
\caption{Summary of optimization strategies across the foundation models considered. Loss function indicates the pre-training objective; forecasting strategy describes how multi-step predictions are generated; multi-patch handling specifies the mechanism used for long-horizon forecasting; and post-training lists additional training stages applied after initial pre-training.}
\label{tab:optim_summary}
\end{table}

%%%%%%%%%%%%%%%%%%%%%%%%%%%%%%%%%%%%%% Transfer Learning: %%%%%%%%%%%%%%%%%%%%%%%%%%%%%%%%%%%%%%%%%%%

\subsection{Transfer Learning via Fine-Tuning on Target Datasets}\label{sec:ajus}

Fine-tuning foundation models is a transfer learning technique that consists in adapting the knowledge acquired during pre-training on large and diverse corpora to improve forecasting performance on a specific target dataset. This is particularly important in the context of time series forecasting, where target datasets are often of limited size and may contain only a small number of time series with short histories, which prevents training expressive deep learning models from scratch. By reusing the temporal representations learned during pre-training across multiple domains, frequencies, and dynamics, fine-tuning makes it possible to overcome this data scarcity constraint.

\begin{figure}[pos = t]
\centering
\includegraphics[width=11cm]{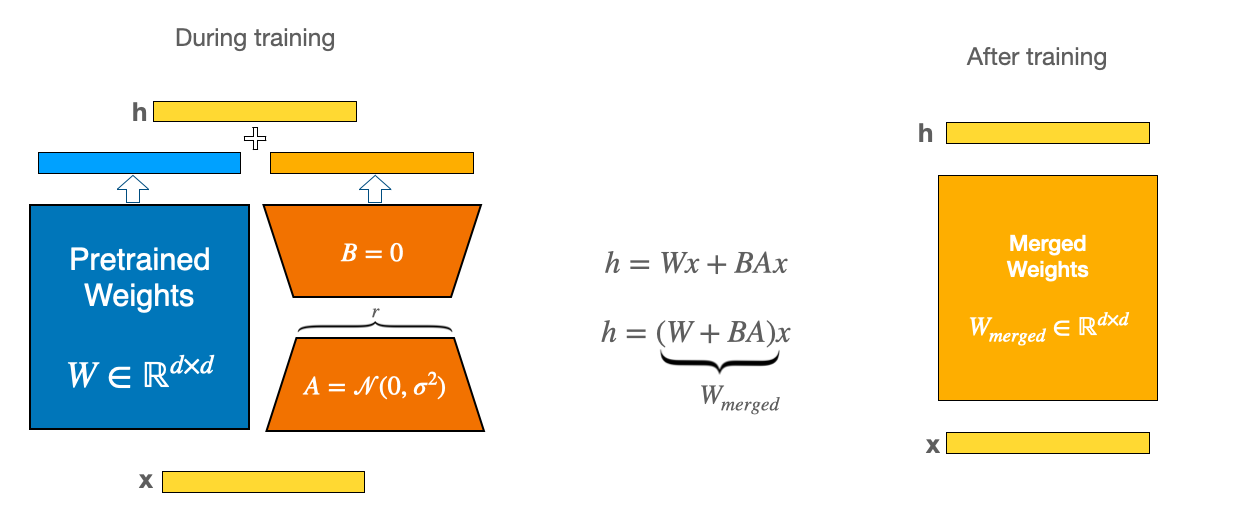}
\caption{Illustration from~\citep{ibm_lora} of Low-Rank Adaptation (LoRA) \citep{hu2022lora}. \textbf{Left (during training):} the pre-trained weight matrix $W \in \mathbb{R}^{d \times d}$ is frozen, and a trainable low-rank perturbation $\Delta W = BA$ is added in parallel, where $A \in \mathbb{R}^{r \times d}$ is initialized from a Gaussian distribution and $B \in \mathbb{R}^{d \times r}$ is initialized to zero, so that the perturbation is initially null. The rank $r \ll d$ controls the number of additional trainable parameters. The output of the adapted layer is $h = Wx + BAx$. \textbf{Right (after training):} the low-rank matrices are merged into the original weights to form $W_{\text{merged}} = W + BA \in \mathbb{R}^{d \times d}$, incurring no additional inference cost compared to the original model.
}
\label{fig:lora}
\end{figure}

In its standard form, referred to as \textit{full fine-tuning}, all model parameters are updated by continuing training on the entire training split of the target dataset. This approach offers maximum flexibility for adaptation but can be computationally expensive and prone to overfitting when the target data is scarce. To address these limitations, two complementary efficiency strategies can be considered:
\begin{itemize}
    \item[(i)] \textbf{Data-efficient fine-tuning} reduces the amount of target data required for adaptation. In the \textit{few-shot} setting, only a small fraction of the target training data (e.g., 20\%) is used to update the model, enabling rapid adaptation with minimal supervision. In the extreme case of the \textit{zero-shot} setting, the pre-trained model is applied directly to the target dataset without any parameter update, relying entirely on the generalization capability acquired during pre-training. This setting serves as a primary benchmark for evaluating foundation models.

    \item[(ii)] \textbf{Parameter-efficient fine-tuning} reduces the number of trainable parameters while keeping most of the model frozen. A prominent method in this category is \textit{Low-Rank Adaptation} (LoRA, \citet{hu2022lora}). The key idea behind LoRA is that the weight updates required for domain adaptation typically lie in a low-dimensional subspace. Rather than modifying the full weight matrix $W \in \mathbb{R}^{d \times k}$ of a given layer, LoRA freezes $W$ and injects a trainable low-rank perturbation of the form $\Delta W = BA$, where $B \in \mathbb{R}^{d \times r}$ and $A \in \mathbb{R}^{r \times k}$ with $r \ll \min(d, k)$. The forward pass of the adapted layer then becomes $h = Wx + BAx$, where $x$ is the input (Figure~\ref{fig:lora}). Since only the matrices $A$ and $B$ are trained, the number of additional parameters is reduced from $d \times k$ to $r \times (d + k)$, which represents a drastic reduction for typical model dimensions. This makes LoRA particularly well suited to fine-tuning large pre-trained models in data-constrained or resource-limited scenarios, while preserving the expressive representations learned during pre-training.
\end{itemize}
These two strategies are complementary and can be combined: for instance, one can apply LoRA in a few-shot setting, achieving both data and parameter efficiency simultaneously. The choice of strategy depends on the available target data, computational resources, and the desired trade-off between adaptation capacity and efficiency.

%%%%%%%%%%%%%%%%%%%%%%%%%%%%%%%%%%%%%% Experiments: %%%%%%%%%%%%%%%%%%%%%%%%%%%%%%%%%%%%%%%%%%%

\section{Experiments}\label{sec:expes}

\subsection{Datasets}
We conduct our fine-tuning study on the full set of datasets constituting the train/test split of the GIFT-Eval benchmark \citep{aksu2024gifteval}, a large-scale evaluation framework dedicated to forecasting models. This portion of the benchmark comprises 15 univariate and 8 multivariate datasets, spanning 7 domains and 10 different frequencies, for a total of \mbox{144{,}000} time series and 177 million observations. In total, 95 unique configurations—combining dataset, frequency, and forecast horizon (short-, medium-, and long-term)—are considered, providing a comprehensive and diverse evaluation of the fine-tuning capabilities of foundation forecasting models.

\subsubsection{Characteristics of Time Series Across Domains}

In this section, we present a detailed analysis of the statistical characteristics of the time series from the train/test datasets of the GIFT-Eval benchmark~\citep{aksu2024gifteval}. This analysis aims to better understand the structural specificities of the data across the different domains considered (finance, healthcare, energy, transport, etc.). To this end, we rely on three fundamental measures commonly used to study the characteristics and predictability of a time series from its history: \textbf{trend}, \textbf{seasonality}, and \textbf{entropy}.

\paragraph{Trend.}
The trend measures the overall evolution of a time series over time (increase, decrease, or stability). It is obtained by decomposing the series into trend, seasonal, and residual components, and then comparing the variance of the residual component to the combined variance of the trend and residual components. The strength of the trend is defined as:
\[
\text{trend} = 1 - \frac{\operatorname{Var}(e_t)}{\operatorname{Var}(f_t + e_t)},
\]
where $f_t$ is the trend component and $e_t$ is the residual. Values close to 1 indicate a pronounced trend, as is often the case for economic or financial series. Conversely, series from domains such as transport tend to exhibit a weaker trend.

\paragraph{Seasonality.}
The strength of seasonality reflects the presence of repetitive patterns at regular intervals—for example, daily cycles (sales, energy) or annual cycles (finance). It is computed for each seasonal component $s_{i,t}$ as:
\[
\text{seasonal\_strength}_i = 1 - \frac{\operatorname{Var}(e_t)}{\operatorname{Var}(s_{i,t} + e_t)}.
\]
A high value indicates the strong presence of recurring patterns, making the series more structured.

\paragraph{Entropy / Predictability.}
Spectral entropy measures the complexity or noise level of a time series:
\[
\text{Entropy} = -\int_{-\pi}^{\pi} \hat{f}(\lambda) \log \hat{f}(\lambda) \, d\lambda,
\]
where $\hat{f}(\lambda)$ is the estimated spectral density. Low entropy reflects a strong underlying structure and an easily predictable series, whereas high entropy characterizes irregular or noisy behavior.
\begin{figure}[pos = t]
    \centering
    \includegraphics[width=7cm]{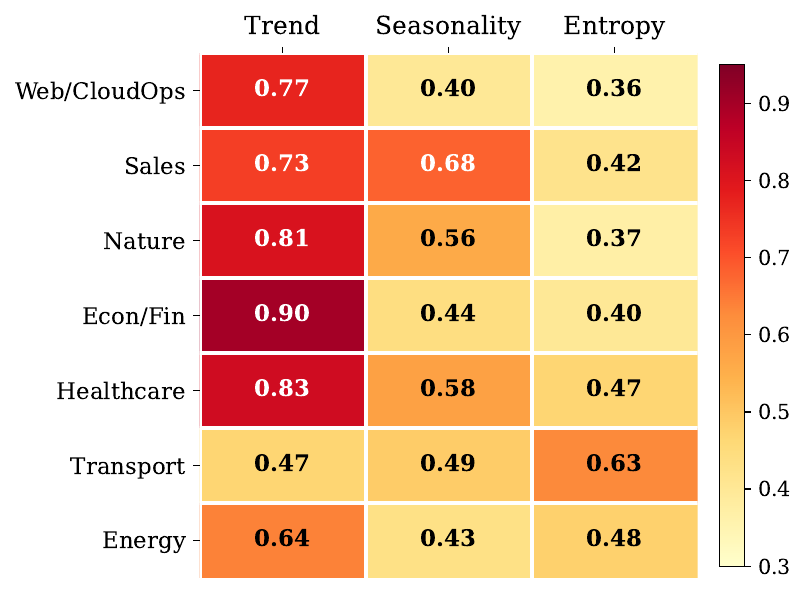}
    \caption{Average trend strength, seasonality strength, and spectral entropy across the seven domains of the GIFT-Eval benchmark. Higher trend and seasonality values indicate more structured and predictable series, while higher entropy reflects greater irregularity. Economic/Financial series exhibit the strongest trend (0.90), whereas Transport stands out with the highest entropy (0.63) and weakest trend (0.47), suggesting more irregular and harder-to-predict dynamics.}
    \label{fig:ts_heatmaps}
\end{figure}

Figure~\ref{fig:ts_heatmaps} shows the average values of these three characteristics for each of the domains considered, highlighting significant structural disparities. These observations confirm the importance of evaluating forecasting models under a diverse set of temporal conditions, as proposed by GIFT-Eval in Figure~\ref{fig:ts_heatmaps}.

\begin{sidewaystable} % <-- HERE
%\begin{table}
\centering
\footnotesize
\setlength{\tabcolsep}{3pt}
\renewcommand{\arraystretch}{1.1}
\resizebox{\linewidth}{!}{%
\begin{tabular}{cccccccccccccccc}
\hline
\textbf{} & \textbf{} & \textbf{} & \textbf{} & \textbf{} 
& \multicolumn{3}{c}{\textbf{Series Length}} 
& \textbf{} & \textbf{} 
& \multicolumn{2}{c}{\textbf{Short-term}} 
& \multicolumn{2}{c}{\textbf{Medium-term}} 
& \multicolumn{2}{c}{\textbf{Long-term}} \\ \hline
\textbf{Dataset} & \textbf{Source} & \textbf{Domain [Size]} & \textbf{Frequency} & \textbf{\# Series} 
& \textbf{Mean} & \textbf{Min} & \textbf{Max} 
& \textbf{\# Obs.} & \textbf{Target Variables} 
& \textbf{Horizon (S)} & \textbf{Test Windows} 
& \textbf{Horizon (M)} & \textbf{Test Windows} 
& \textbf{Horizon (L)} & \textbf{Test Windows} \\ \hline
Jena Weather & Autoformer~\citep{Wu2021AutoformerDT} & Nature [Large]  & 10T & 1 & 52,704 & 52,704 & 52,704 & 52,704 & 21 & 48 & 20 & 480 & 11 & 720 & 8 \\
Jena Weather & Autoformer~\citep{Wu2021AutoformerDT} & Nature [Medium] & H   & 1 & 8,784  & 8,784  & 8,784  & 8,784  & 21 & 48 & 19 & 480 & 2  & 720 & 2 \\
Jena Weather & Autoformer~\citep{Wu2021AutoformerDT} & Nature [Small]  & D   & 1 & 366    & 366    & 366    & 366    & 21 & 30 & 2  &     &    &     &   \\
BizITObs - Application & AutoMixer~\citep{10.1609/aaai.v38i21.30336} & Web/CloudOps [Small]  & 10S & 1  & 8,834  & 8,834  & 8,834  & 8,834    & 2 & 60 & 15 & 600 & 2 & 900 & 1 \\
BizITObs - Service     & AutoMixer~\citep{10.1609/aaai.v38i21.30336} & Web/CloudOps [Medium] & 10S & 21 & 8,835  & 8,835  & 8,835  & 185,535  & 2 & 60 & 15 & 600 & 2 & 900 & 1 \\
BizITObs - L2C         & AutoMixer~\citep{10.1609/aaai.v38i21.30336} & Web/CloudOps [Medium] & 5T  & 1  & 31,968 & 31,968 & 31,968 & 31,968   & 7 & 48 & 20 & 480 & 7 & 720 & 5 \\
BizITObs - L2C         & AutoMixer~\citep{10.1609/aaai.v38i21.30336} & Web/CloudOps [Small]  & H   & 1  & 2,664  & 2,664  & 2,664  & 2,664    & 7 & 48 & 6  & 480 & 1 & 720 & 1 \\
Bitbrains - Fast Storage & Grid Workloads Archive~\citep{10.1109/CCGrid.2015.60} & Web/CloudOps [Large] & 5T & 1,250 & 8,640 & 8,640 & 8,640 & 10,800,000 & 2 & 48 & 18 & 480 & 2 & 720 & 2 \\
Bitbrains - Fast Storage & Grid Workloads Archive~\citep{10.1109/CCGrid.2015.60} & Web/CloudOps [Large] & H  & 1,250 & 721   & 721   & 721   & 901,250    & 2 & 48 & 2  &     &   &     &   \\
Bitbrains - rnd          & Grid Workloads Archive~\citep{10.1109/CCGrid.2015.60} & Web/CloudOps [Large] & 5T & 500   & 8,640 & 8,640 & 8,640 & 4,320,000  & 2 & 48 & 18 & 480 & 2 & 720 & 2 \\
Bitbrains - rnd          & Grid Workloads Archive~\citep{10.1109/CCGrid.2015.60} & Web/CloudOps [Large] & H  & 500   & 720   & 720   & 720   & 360,000    & 2 & 48 & 2  &     &   &     &   \\
ETT1 & Informer~\citep{Zhou2020InformerBE} & Energy [Medium] & 15T   & 1 & 69,680 & 69,680 & 69,680 & 69,680 & 7 & 48 & 20 & 480 & 15 & 720 & 10 \\
ETT1 & Informer~\citep{Zhou2020InformerBE} & Energy [Medium] & H     & 1 & 17,420 & 17,420 & 17,420 & 17,420 & 7 & 48 & 20 & 480 & 4  & 720 & 3  \\
ETT1 & Informer~\citep{Zhou2020InformerBE} & Energy [Small]  & D     & 1 & 725    & 725    & 725    & 725    & 7 & 30 & 3  &     &    &     &    \\
ETT1 & Informer~\citep{Zhou2020InformerBE} & Energy [Small]  & W-THU & 1 & 103    & 103    & 103    & 103    & 7 & 8  & 2  &     &    &     &    \\
ETT2 & Informer~\citep{Zhou2020InformerBE} & Energy [Medium] & 15T   & 1 & 69,680 & 69,680 & 69,680 & 69,680 & 7 & 48 & 20 & 480 & 15 & 720 & 10 \\
ETT2 & Informer~\citep{Zhou2020InformerBE} & Energy [Medium] & H     & 1 & 17,420 & 17,420 & 17,420 & 17,420 & 7 & 48 & 20 & 480 & 4  & 720 & 3  \\
ETT2 & Informer~\citep{Zhou2020InformerBE} & Energy [Small]  & D     & 1 & 725    & 725    & 725    & 725    & 7 & 30 & 3  &     &    &     &    \\
ETT2 & Informer~\citep{Zhou2020InformerBE} & Energy [Small]  & W-THU & 1 & 103    & 103    & 103    & 103    & 7 & 8  & 2  &     &    &     &    \\
Loop Seattle & LibCity~\citep{Wang2023TowardsEA} & Transport [Large] & 5T & 323 & 105,120 & 105,120 & 105,120 & 33,953,760 & 1 & 48 & 20 & 480 & 20 & 720 & 15 \\
Loop Seattle & LibCity~\citep{Wang2023TowardsEA} & Transport [Large] & H  & 323 & 8,760   & 8,760   & 8,760   & 2,829,480  & 1 & 48 & 19 & 480 & 2  & 720 & 2  \\
Loop Seattle & LibCity~\citep{Wang2023TowardsEA} & Transport [Small] & D  & 323 & 365     & 365     & 365     & 117,895    & 1 & 30 & 2  &     &    &     &    \\
SZ-Taxi      & LibCity~\citep{Wang2023TowardsEA} & Transport [Medium] & 15T & 156 & 2,976 & 2,976 & 2,976 & 464,256 & 1 & 48 & 7 & 480 & 1 & 720 & 1 \\
SZ-Taxi      & LibCity~\citep{Wang2023TowardsEA} & Transport [Medium] & H   & 156 & 744   & 744   & 744   & 116,064 & 1 & 48 & 2 &     &   &     &   \\
M\_DENSE     & LibCity~\citep{Wang2023TowardsEA} & Transport [Large] & H & 30 & 17,520 & 17,520 & 17,520 & 525,600 & 1 & 48 & 20 & 480 & 4 & 720 & 3 \\
M\_DENSE     & LibCity~\citep{Wang2023TowardsEA} & Transport [Small] & D & 30 & 730    & 730    & 730    & 21,900  & 1 & 30 & 3  &     &   &     &   \\
Solar & LSTNet~\citep{Lai2017ModelingLA} & Energy [Large] & 10T   & 137 & 52,560 & 52,560 & 52,560 & 7,200,720 & 1 & 48 & 20 & 480 & 11 & 720 & 8 \\
Solar & LSTNet~\citep{Lai2017ModelingLA} & Energy [Large] & H     & 137 & 8,760  & 8,760  & 8,760  & 1,200,120 & 1 & 48 & 19 & 480 & 2  & 720 & 2 \\
Solar & LSTNet~\citep{Lai2017ModelingLA} & Energy [Small] & D     & 137 & 365    & 365    & 365    & 50,005    & 1 & 30 & 2  &     &    &     &   \\
Solar & LSTNet~\citep{Lai2017ModelingLA} & Energy [Small] & W-FRI & 137 & 52     & 52     & 52     & 7,124     & 1 & 8  & 1  &     &    &     &   \\
Hierarchical Sales & \citet{Mancuso2020AML} & Sales [Medium] & D     & 118 & 1,825 & 1,825 & 1,825 & 215,350 & 1 & 30 & 7 & & & & \\
Hierarchical Sales & \citet{Mancuso2020AML} & Sales [Small]  & W-WED & 118 & 260   & 260   & 260   & 30,680  & 1 & 8  & 4 & & & & \\
M4 Monthly & Monash~\citep{godahewa2021monash} & Econ/Fin [Large]  & M     & 48,000 & 234   & 60  & 2,812 & 11,246,411 & 1 & 18 & 1 & & & & \\
M4 Weekly  & Monash~\citep{godahewa2021monash} & Econ/Fin [Medium] & W-SUN & 359    & 1,035 & 93  & 2,610 & 371,579    & 1 & 13 & 1 & & & & \\
M4 Daily   & Monash~\citep{godahewa2021monash} & Econ/Fin [Large]  & D     & 4,227  & 2,371 & 107 & 9,933 & 10,023,836 & 1 & 14 & 1 & & & & \\
M4 Hourly  & Monash~\citep{godahewa2021monash} & Econ/Fin [Medium] & H     & 414    & 902   & 748 & 1,008 & 373,372    & 1 & 48 & 2 & & & & \\
Hospital      & Monash~\citep{godahewa2021monash} & Healthcare [Medium] & M & 767 & 84  & 84  & 84  & 64,428 & 1 & 12 & 1  & & & & \\
COVID Deaths  & Monash~\citep{godahewa2021monash} & Healthcare [Medium] & D & 266 & 212 & 212 & 212 & 56,392 & 1 & 30 & 1  & & & & \\
US Births & Monash~\citep{godahewa2021monash} & Healthcare [Small] & D     & 1 & 7,305 & 7,305 & 7,305 & 7,305 & 1 & 30 & 20 & & & & \\
US Births & Monash~\citep{godahewa2021monash} & Healthcare [Small] & W-TUE & 1 & 1,043 & 1,043 & 1,043 & 1,043 & 1 & 8  & 14 & & & & \\
US Births & Monash~\citep{godahewa2021monash} & Healthcare [Small] & M     & 1 & 240   & 240   & 240   & 240   & 1 & 12 & 2  & & & & \\
Saugeen & Monash~\citep{godahewa2021monash} & Nature [Small] & D     & 1 & 23,741 & 23,741 & 23,741 & 23,741 & 1 & 30 & 20 & & & & \\
Saugeen & Monash~\citep{godahewa2021monash} & Nature [Small] & W-THU & 1 & 3,391  & 3,391  & 3,391  & 3,391  & 1 & 8  & 20 & & & & \\
Saugeen & Monash~\citep{godahewa2021monash} & Nature [Small] & M     & 1 & 780    & 780    & 780    &        & 1 & 12 & 7  & & & & \\
Temperature Rain & Monash~\citep{godahewa2021monash} & Nature [Large]  & D & 32,072 & 725    & 725   & 725    & 780       & 1 & 30 & 3  & & & & \\
KDD Cup 2018     & Monash~\citep{godahewa2021monash} & Nature [Large]  & H & 270    & 10,898 & 9,504 & 10,920 & 2,942,364 & 1 & 48 & 20 & 480 & 2 & 720 & 2 \\
KDD Cup 2018     & Monash~\citep{godahewa2021monash} & Nature [Medium] & D & 270    & 455    & 396   & 455    & 122,791   & 1 & 30 & 2  &     &   &     &   \\
Car Parts & Monash~\citep{godahewa2021monash} & Sales [Medium] & M & 2,674 & 51 & 51 & 51 & 136,374 & 1 & 12 & 1 & & & & \\
Electricity & UCI ML Archive~\citep{electricityloaddiagrams20112014_321} & Energy [Large]  & 15T   & 370 & 140,256 & 140,256 & 140,256 & 51,894,720 & 1 & 48 & 20 & 480 & 20 & 720 & 20 \\
Electricity & UCI ML Archive~\citep{electricityloaddiagrams20112014_321} & Energy [Large]  & H     & 370 & 35,064  & 35,064  & 35,064  & 12,973,680 & 1 & 48 & 20 & 480 & 8  & 720 & 5  \\
Electricity & UCI ML Archive~\citep{electricityloaddiagrams20112014_321} & Energy [Large]  & D     & 370 & 1,461   & 1,461   & 1,461   & 540,570    & 1 & 30 & 5  &     &    &     &    \\
Electricity & UCI ML Archive~\citep{electricityloaddiagrams20112014_321} & Energy [Medium] & W-FRI & 370 & 208     & 208     & 208     & 76,960     & 1 & 8  & 3  &     &    &     &    \\ \hline
\end{tabular}%
}
\caption{Individual statistics for each dataset in the benchmark, including the size of each one (source:~\cite{aksu2024gifteval}).}
\label{tab:dataset}
%\end{table}
\end{sidewaystable} % <-- HERE

\clearpage

\subsubsection{Datasets Used and Their Sizes}

Table~\ref{tab:dataset} presents the complete set of datasets used in our study, along with their main characteristics, including size, domain, frequency, and average series length. This information provides a clearer picture of the diversity and complexity of the data on which the models are evaluated. Sources of the datasets can be found in 
\citep{Wu2021AutoformerDT,10.1609/aaai.v38i21.30336, 10.1109/CCGrid.2015.60,Zhou2020InformerBE, Wang2023TowardsEA, Lai2017ModelingLA, Mancuso2020AML, godahewa2021monash, electricityloaddiagrams20112014_321}.

\subsection{Experimental Models}

We focus in this study on the top-performing models from the pre-trained section of the GIFT-Eval leaderboard\footnote{\url{https://huggingface.co/spaces/Salesforce/GIFT-Eval}}, on April 9th 2026, the date at which we froze the model selection. Among the top five models at that time, only two--Chronos~2, ranked first, and TTM-R3-PT, ranked fifth--were publicly available for fine-tuning purposes, thus allowing full replication in this setting, and had no test leakage, thus enabling accurate evaluation. 
%which have no test lea   dataset The choice of models is based on two criteria: their performance (MWQL) on the GIFT-Eval benchmark and the availability of their training code. Among the models with publicly available training code, we selected the two top-performing modelR3-PT s  the GIFT-Eval leaderboard\footnote{\url{https://huggingface.co/spaces/Salesforce/GIFT-Eval}, accessed: 09/04/2026.} at the time of our study: Chronos~2 and TTM-R3-PT.
We compare the performance of the fine-tuned versions of these models (see Section~\ref{sec:tuning} below) with that of their zero-shot counterparts. It is important to note, however, that Chronos~2 and TTM-R3-PT were pre-trained on a subset of the GIFT-Eval training data. Consequently, for some datasets in our benchmark, the zero-shot evaluation of these models is not entirely rigorous, as they may have been exposed to part of the training data during pre-training, which shares common properties with the test sets. Nevertheless, the impact of this contamination remains limited in our setting, since we compare each zero-shot model only against its own fine-tuned version, both sharing the same pre-training corpus. This issue becomes more significant, however, when comparing different models against one another in the zero-shot setting, as emphasized by \citet{aksu2024gifteval}.

\subsection{Experimental Protocol}

Given the large number of experiments to be conducted, we do not perform a search for optimal hyperparameters for each dataset; instead, we use the same hyperparameters for a given model across all datasets. Table~\ref{tab:params} reports the hyperparameters used for fine-tuning each model, for which we retain the default values. The study of optimal hyperparameter selection is left for future work.

\paragraph{Fine-tuning details.}\label{sec:tuning}
We fine-tune both Chronos~2 and TTM-R3-PT on each target dataset, updating all model parameters using the entire training split (full-shot setting, also referred to as \textit{full-fine tuning}), with a fixed computational budget of $4{,}000$ gradient update steps. We do not consider few-shot fine-tuning in this setting, as updating all parameters with limited data is prone to overfitting. Additionally, we explore parameter-efficient fine-tuning via LoRA, under both full-shot and few-shot (20\% of the training data) settings, in order to assess the trade-off between adaptation capacity and data efficiency. For this purpose, we used the AutoGluon library~\citep{agtimeseries} for Chronos~2 and the official code repository for TTM\footnote{\url{https://github.com/ibm-granite/granite-tsfm/tree/ttm-r3-release-mq2}}.

\begin{table}[pos = t]
\begin{center}
\begin{tabular}{lcc}
\hline
\textbf{Parameter} & \textbf{Chronos~2} & \textbf{TTM-R3-PT} \\
\hline
Learning rate (full fine-tuning) & $1\times10^{-4}$ & $1\times10^{-3}$ \\
Learning rate (LoRA) & $1\times10^{-5}$ & $1\times10^{-5}$ \\
Batch size & 32 & 18--256\textsuperscript{$\dagger$} \\
Max.\ gradient steps & 4{,}000 & 4{,}000 \\
Optimized objective & Quantile loss & Quantile loss \\
LoRA rank ($r$) & 8 & 8 \\
LoRA alpha ($\alpha$) & 16 & 16 \\
\hline
\multicolumn{3}{l}{\small \textsuperscript{$\dagger$}Adapted to dataset size.} \\
\end{tabular}
\caption{Hyperparameters used for fine-tuning. LoRA-specific parameters (rank, alpha, and learning rate) apply only to the parameter-efficient fine-tuning setting.}
\label{tab:params}
\end{center}
\end{table}

\paragraph{Evaluation Metrics.}
We assess the performance of the forecasting models using two complementary metrics: the \textit{Mean Absolute Percentage Error} (MAPE) for point forecasting, and the \textit{Continuous Ranked Probability Score} (CRPS) for probabilistic forecasting.

The MAPE measures the average absolute percentage difference between the predicted values $\hat{y}_t$ (we use the median prediction) and the ground-truth values $y_t$, and is defined as:
\begin{equation}
\text{MAPE} = \frac{100}{H} \sum_{t=T+1}^{T+H} \left| \frac{y_t - \hat{y}_t}{y_t} \right|,
\end{equation}
where $H$ denotes the forecast horizon.

The CRPS is a metric used in probabilistic forecasting to evaluate the accuracy of predicted cumulative distribution functions $F$ with respect to observed values. For a ground-truth value $y$, the CRPS is defined as:
\begin{equation}
\text{CRPS}(F, y) = \int_0^1 2 \, \Lambda_\alpha(F^{-1}(\alpha), y) \, d\alpha,
\end{equation}
where the quantile loss $\Lambda_\alpha(q, y)$ is given by $\Lambda_\alpha(q, y) = (\alpha - \mathbf{1}_{\{y < q\}})(y - q)$, with $\mathbf{1}_{\{y < q\}}$ the indicator function equal to 1 if $y < q$, and 0 otherwise.

\medskip

In practice, computing the CRPS integral can be computationally expensive, and we approximate it with a discrete sum over a finite set of quantile levels. This approximation, commonly referred to as the \textit{Mean Weighted Quantile Loss} (MWQL, \cite{aksu2024gifteval}), is defined as:
\[
\mathrm{MWQL} = \frac{2}{K} \sum_{k=1}^{K} \frac{\sum_{t=T+1}^{T+H} \Lambda_{\alpha_k}(\hat{q}_t(\alpha_k), y_t)}{\sum_{t=T+1}^{T+H} |y_t|},
\]
where $K$ is the number of quantile levels, $Q = \{\alpha_1, \alpha_2, \ldots, \alpha_K\}$ is the set of selected quantiles with $\alpha_k = 0.1k$ for $k \in \{1, 2, \ldots, 9\}$ when $K = 9$, $\hat{q}_t(\alpha)$ is the predicted quantile at level $\alpha$ and time $t$, $y_t$ is the ground-truth observed value, and $\Lambda_\alpha(\hat{q}_t(\alpha), y_t)$ is the quantile loss, which captures the discrepancy between the predicted quantile and the observation through an asymmetric penalization dependent on $\alpha$.

It is worth noting that the quantile loss and the mixture-distribution likelihood (used for training the models considered in this study, see Table~\ref{tab:params}) are directly related to the $\mathrm{MWQL}$. Optimizing these objectives thus translates into a reduction of the $\mathrm{MWQL}$, improving the quality of probabilistic forecasts.

Since the magnitude of the evaluation metrics can vary substantially across datasets, we normalize each model's metrics by dividing them by those of a reference model (here, Seasonal Naive): $\frac{\mathrm{Metric}_{\text{model}}}{\mathrm{Metric}_{\text{ref}}}$. The resulting relative scores are then aggregated using the arithmetic mean. We denote these normalized metrics by $\mathrm{MWQL}_n$ and $\mathrm{MAPE}_n$.

\subsection{Results and Analysis}

We organize our analysis along two complementary axes. First, we aggregate performance by \textbf{dataset size}---small, medium, and large, defined using the 0.33 and 0.66 quantiles of the number of observations per dataset---to examine how the volume of available training data affects fine-tuning quality (Table~\ref{tab:size_aggregated}). Second, we provide a \textbf{domain-level} analysis, disaggregating results by application domain and forecast horizon (short, medium, and long-term, as defined in \citet{aksu2024gifteval}) to identify the conditions under which each adaptation strategy is most effective (Table~\ref{tab:domain_aggregated}).

%%%%%%%%%%%%%%%%%%%%%%%%%%%%%%%%%% Table 3 %%%%%%%%%%%%%%%%%%%%%%%%%%%%%%%%%%%%%%%%%%%%%%
\begin{table}[pos = t]
 \centering
 \small
 \resizebox{\textwidth}{!}{
 \begin{tabular}{ll|cccc|cccc}
 \hline
 \textbf{Size} & \textbf{Metric} & \multicolumn{4}{c|}{\textbf{Chronos2}} & \multicolumn{4}{c}{\textbf{TTM-R3-PT}} \\
 \cline{3-10}
  &  & Zero-Shot & FFT & LoRA-FullShot & LoRA-FewShot & Zero-Shot & FFT & LoRA-FullShot & LoRA-FewShot \\
 \hline
 \multirow{2}{*}{Small} & $\mathrm{MWQL}_n$ & \textbf{0.423} & 0.577±0.000 & 0.477±0.006 & 0.471±0.007 & 0.494 & \textbf{0.492±0.002} & 0.531±0.002 & 0.533±0.004 \\
  & $\mathrm{MAPE}_n$ & \textbf{0.544} & 0.732±0.000 & 0.647±0.002 & 0.612±0.006 & \textbf{0.603} & 0.613±0.002 & 0.638±0.008 & 0.637±0.005 \\
 \hline
 \multirow{2}{*}{Medium} & $\mathrm{MWQL}_n$ & 0.364 & 0.387±0.001 & \textbf{0.357±0.001} & 0.394±0.001 & \textbf{0.391} & \textbf{0.391±0.001} & 0.398±0.001 & 0.402±0.002 \\
  & $\mathrm{MAPE}_n$ & \textbf{0.703} & 0.824±0.000 & 0.722±0.002 & 0.796±0.003 & \textbf{0.738} & 0.740±0.003 & 0.743±0.004 & 0.756±0.002 \\
 \hline
 \multirow{2}{*}{Large} & $\mathrm{MWQL}_n$ & 0.310 & 0.322±0.000 & \textbf{0.301±0.002} & 0.330±0.003 & 0.355 & \textbf{0.350±0.000} & 0.352±0.000 & 0.351±0.000 \\
  & $\mathrm{MAPE}_n$ & 1.006 & 1.266±0.002 & 1.056±0.035 & \textbf{0.931±0.016} & 1.240 & \textbf{1.219±0.005} & 1.248±0.015 & 1.254±0.011 \\
 \hline
 \end{tabular}
 }
 \caption{Average $\mathrm{MWQL}_n$ and $\mathrm{MAPE}_n$ within each dataset size category for Chronos2 and TTM-R3-PT under Zero-Shot, full fine-tuning (FFT), and LoRA-based adaptation strategies (LoRA-FullShot and LoRA-FewShot). Results for FFT and LoRA are reported as mean±std over three seeds. Bold indicates the best result per metric and dataset size across all adaptation strategies.}
 \label{tab:size_aggregated}
 \end{table}

%%%%%%%%%%%%%%%%%%%%%%%%%%%%%%%%%% Table 4 %%%%%%%%%%%%%%%%%%%%%%%%%%%%%%%%%%%%%%%%%%%%%%

\begin{table}[pos=t]
\centering
\small
\resizebox{\textwidth}{!}{
\begin{tabular}{ll|cccc|cccc}
\hline
& &\multicolumn{4}{c|}{\textbf{Chronos2}} & \multicolumn{4}{c}{\textbf{TTM-R3-PT}} \\
\cline{3-10}
\textbf{Domain} & \textbf{Horizon} & Zero-shot & FFT & LoRA-FullShot & LoRA-FewShot & Zero-shot & FFT & LoRA-FullShot & LoRA-FewShot \\
\hline
&& \multicolumn{8}{c}{\textbf{$\mathrm{MWQL}_n$}} \\
\hline
\multirow{3}{*}{Transport} & Short & 0.314 & \textbf{0.311±0.000} & 0.314±0.001 & 0.339±0.003 & \textbf{0.336} & 0.338±0.000 & 0.339±0.002 & 0.337±0.001 \\
 & Medium & 0.146 & \textbf{0.138±0.000} & 0.140±0.000 & 0.144±0.000 & 0.143 & \textbf{0.142±0.000} & \textbf{0.142±0.000} & \textbf{0.142±0.001} \\
 & Long & 0.123 & \textbf{0.120±0.000} & 0.127±0.000 & 0.126±0.000 & 0.125 & \textbf{0.124±0.000} & \textbf{0.124±0.000} & \textbf{0.124±0.000} \\
\hline
\multirow{3}{*}{Nature} & Short & \textbf{0.367} & 0.435±0.000 & 0.381±0.000 & 0.388±0.006 & 0.419 & 0.415±0.001 & 0.412±0.001 & \textbf{0.410±0.003} \\
 & Medium & \textbf{0.174} & 0.180±0.000 & 0.175±0.000 & 0.179±0.000 & 0.188 & 0.187±0.000 & \textbf{0.186±0.000} & 0.187±0.001 \\
 & Long & 0.153 & 0.161±0.000 & \textbf{0.152±0.000} & 0.153±0.001 & 0.162 & 0.162±0.000 & \textbf{0.161±0.001} & \textbf{0.161±0.001} \\
\hline
\multirow{3}{*}{Web/CloudOps} & Short & 0.469 & 0.572±0.001 & \textbf{0.460±0.017} & 0.470±0.000 & \textbf{0.536} & 0.554±0.004 & 0.547±0.006 & 0.541±0.008 \\
 & Medium & 0.382 & 0.464±0.000 & \textbf{0.380±0.010} & 0.433±0.006 & \textbf{0.399} & 0.406±0.004 & 0.405±0.006 & 0.411±0.002 \\
 & Long & 0.393 & 0.435±0.000 & \textbf{0.383±0.006} & 0.504±0.015 & 0.420 & 0.410±0.002 & 0.409±0.001 & \textbf{0.405±0.001} \\
\hline
\multirow{1}{*}{Econ/Fin} & Short & 0.603 & 0.615±0.001 & \textbf{0.601±0.001} & 0.606±0.000 & 0.746 & \textbf{0.720±0.000} & 0.736±0.003 & 0.729±0.003 \\

\hline
\multirow{3}{*}{Energy} & Short & \textbf{0.520} & 0.617±0.001 & 0.561±0.002 & 0.544±0.003 & 0.556 & \textbf{0.550±0.003} & 0.565±0.009 & 0.571±0.009 \\
 & Medium & \textbf{0.255} & 0.278±0.000 & 0.259±0.000 & 0.302±0.002 & 0.271 & \textbf{0.270±0.001} & \textbf{0.270±0.002} & 0.271±0.002 \\
 & Long & \textbf{0.225} & 0.251±0.000 & 0.227±0.000 & 0.272±0.001 & \textbf{0.234} & 0.236±0.001 & \textbf{0.234±0.001} & \textbf{0.234±0.001} \\
\hline
\multirow{1}{*}{Healthcare} & Short & 0.360 & 0.439±0.000 & \textbf{0.341±0.005} & 0.405±0.011 & 0.450 & \textbf{0.420±0.003} & 0.593±0.022 & 0.612±0.012 \\

\hline
\multirow{1}{*}{Sales} & Short & 0.369 & 0.381±0.000 & \textbf{0.364±0.000} & 0.379±0.004 & \textbf{0.407} & 0.412±0.004 & 0.441±0.009 & 0.453±0.010 \\
\hline
&& \multicolumn{8}{c}{\textbf{$\mathrm{MAPE}_n$}} \\
\hline
\multirow{3}{*}{Transport} & Short & 0.591 & \textbf{0.576±0.000} & 0.600±0.001 & 0.619±0.005 & 0.635 & 0.623±0.003 & 0.618±0.004 & \textbf{0.616±0.006} \\
 & Medium & 0.517 & 0.498±0.000 & 0.532±0.003 & \textbf{0.477±0.001} & \textbf{0.486} & 0.488±0.002 & 0.487±0.001 & \textbf{0.486±0.001} \\
 & Long & 0.476 & 0.405±0.000 & \textbf{0.400±0.011} & 0.435±0.001 & 0.437 & \textbf{0.426±0.001} & 0.431±0.001 & 0.431±0.001 \\
\hline
\multirow{3}{*}{Nature} & Short & \textbf{0.677} & 0.939±0.000 & 0.704±0.005 & 0.760±0.027 & 0.719 & \textbf{0.692±0.003} & 0.698±0.001 & 0.712±0.002 \\
 & Medium & \textbf{0.901} & 1.295±0.000 & 1.023±0.010 & 1.297±0.052 & \textbf{0.813} & 0.862±0.019 & 0.863±0.017 & 0.850±0.005 \\
 & Long & \textbf{1.042} & 1.549±0.000 & 1.174±0.011 & 1.147±0.040 & 1.015 & 1.014±0.005 & \textbf{1.007±0.019} & 1.036±0.011 \\
\hline
\multirow{3}{*}{Web/CloudOps} & Short & \textbf{0.511} & 0.698±0.003 & 0.531±0.032 & 0.523±0.006 & \textbf{0.576} & 0.604±0.008 & 0.676±0.008 & 0.658±0.026 \\
 & Medium & \textbf{0.690} & 1.146±0.000 & 0.859±0.147 & 0.791±0.005 & 1.264 & \textbf{0.928±0.041} & 1.026±0.099 & 1.107±0.035 \\
 & Long & 0.695 & \textbf{0.651±0.000} & 0.737±0.014 & 0.812±0.029 & 0.995 & 0.880±0.003 & 0.880±0.009 & \textbf{0.870±0.015} \\
\hline
\multirow{1}{*}{Econ/Fin} & Short & \textbf{0.663} & 0.696±0.001 & 0.664±0.001 & 0.681±0.000 & \textbf{0.811} & 0.813±0.002 & 0.876±0.005 & 0.866±0.005 \\

\hline
\multirow{3}{*}{Energy} & Short & 0.939 & 1.158±0.008 & 1.049±0.005 & \textbf{0.916±0.000} & \textbf{1.086} & 1.123±0.004 & 1.117±0.006 & 1.121±0.007 \\
 & Medium & 1.012 & 1.137±0.000 & 0.969±0.010 & \textbf{0.870±0.014} & \textbf{1.115} & 1.153±0.007 & 1.142±0.001 & 1.138±0.003 \\
 & Long & 1.457 & 2.029±0.000 & 1.601±0.007 & \textbf{1.364±0.009} & \textbf{1.763} & 1.902±0.019 & 1.857±0.008 & 1.866±0.009 \\
\hline
\multirow{1}{*}{Healthcare} & Short & 0.500 & 0.545±0.000 & \textbf{0.483±0.008} & 0.558±0.016 & 0.578 & \textbf{0.538±0.002} & 0.680±0.017 & 0.697±0.021 \\

\hline
\multirow{1}{*}{Sales} & Short & 0.681 & 0.762±0.000 & 0.709±0.001 & \textbf{0.673±0.004} & \textbf{0.708} & 0.771±0.011 & 0.801±0.004 & 0.829±0.018 \\
\hline
\end{tabular}
}
\caption{Average $\mathrm{MWQL}_n$ and $\mathrm{MAPE}_n$ within each domain and forecast horizon for Chronos2 and TTM-R3-PT under Zero-Shot, full fine-tuning (FFT), and LoRA-based adaptation strategies (LoRA-FullShot and LoRA-FewShot). Results for FFT and LoRA are reported as mean$\pm$std over three seeds. Bold indicates the best result per metric, domain, and horizon across all adaptation strategies.}
\label{tab:domain_aggregated}
\end{table}

%%%%%%%%%%%%%%%%%%%%%%%%%%%%%%%%%%% Analysis and Results %%%%%%%%%%%%%%%%%%%%%%%%%%%%%%%%%%%%%%%%

\subsubsection{Effect of Dataset Size}
 
As shown in Table~\ref{tab:size_aggregated}, dataset size has a pronounced effect on the relative benefit of fine-tuning, and the two models exhibit markedly different behaviors.
 
For \textbf{Chronos~2}, zero-shot performance is strongest on small datasets ($\mathrm{MWQL}_n = 0.423$, $\mathrm{MAPE}_n = 0.544$), where all fine-tuning strategies degrade performance---full fine-tuning (FFT) in particular, which increases $\mathrm{MWQL}_n$ to $0.577$. This suggests that small datasets provide insufficient data to adapt Chronos~2's 120M parameters without overfitting, even with LoRA. On medium and large datasets, however, LoRA-FullShot becomes the best strategy, achieving $\mathrm{MWQL}_n = 0.357$ and $0.301$ respectively, improving over zero-shot by 2\% and 3\%. Notably, FFT consistently underperforms LoRA on Chronos~2 across all size categories, indicating that parameter-efficient adaptation is better suited to this model's scale.
 
For \textbf{TTM-R3-PT}, FFT provides modest improvements on large datasets ($\mathrm{MWQL}_n$: $0.350$ vs.\ $0.355$ at zero-shot), while LoRA variants offer little to no gain and occasionally degrade performance. This is consistent with TTM's compact architecture (1M to 36M parameters), which is already well-suited to direct weight updates and does not benefit from the additional regularization that LoRA provides to larger models.

\subsubsection{Domain and Horizon Analysis}
 
Table~\ref{tab:domain_aggregated} reveals substantial variation across domains and horizons, highlighting that no single strategy dominates universally.
 
\paragraph{Transport.} Fine-tuning is generally beneficial for both models. For Chronos~2, FFT achieves the best $\mathrm{MWQL}_n$ across all three horizons (e.g., $0.138$ vs.\ $0.146$ at medium-term, a 5.5\% improvement over zero-shot). For TTM, FFT and LoRA yield similar marginal improvements over zero-shot. On $\mathrm{MAPE}_n$, LoRA (both few-shot and full-shot) achieves the best result in four out of six settings across both models (e.g., $0.477$ for LoRA-FewShot vs.\ $0.498$ for FFT on Chronos~2; $0.616$ for LoRA-FewShot vs.\ $0.623$ for FFT on TTM-R3-PT), suggesting that lighter adaptation is sufficient on high-entropy datasets characterized by noise and irregular patterns, helping to avoid overfitting to noisy dynamics.
 
\paragraph{Nature.} Zero-shot Chronos~2 is remarkably strong on this domain, achieving the best $\mathrm{MWQL}_n$ at short-term ($0.367$) and medium-term ($0.174$). FFT consistently degrades performance, particularly on $\mathrm{MAPE}_n$, where it inflates error by 39\% at short-term ($0.939$ vs.\ $0.677$), likely due to severe overfitting when fully updating all 120M parameters. For TTM, LoRA provides small but consistent gains across all horizons on $\mathrm{MWQL}_n$.
 
\paragraph{Web/CloudOps.} LoRA-FullShot proves to be the most effective strategy for Chronos~2, achieving the best $\mathrm{MWQL}_n$ at all three horizons ($0.460$, $0.380$, $0.383$). FFT again degrades performance substantially. For TTM, zero-shot remains competitive at short- and medium-term, while LoRA-FewShot achieves the best long-term $\mathrm{MWQL}_n$ ($0.405$).
 
\paragraph{Energy.} Zero-shot Chronos~2 dominates on $\mathrm{MWQL}_n$ across all horizons ($0.520$, $0.255$, $0.225$). However, on $\mathrm{MAPE}_n$, LoRA-FewShot consistently outperforms all other strategies, achieving the best scores at all three horizons ($0.916$, $0.870$, $1.364$). This discrepancy between the two metrics suggests that LoRA-FewShot 
improves point forecast accuracy without improving overall probabilistic 
calibration, indicating that fine-tuning may sharpen the median prediction 
at the cost of a less well-calibrated predictive distribution. For TTM, FFT provides minor improvements at short-term, while LoRA variants match zero-shot at medium and long-term.
 
\paragraph{Healthcare, Sales, and Econ/Fin.} These domains, each represented by a single horizon, further confirm the general trends. LoRA-FullShot is the best Chronos~2 strategy for Healthcare ($\mathrm{MWQL}_n = 0.341$) and Sales ($0.364$), while zero-shot remains strongest for Econ/Fin on $\mathrm{MAPE}_n$ ($0.663$). For TTM, FFT is consistently the best or near-best strategy across these domains, while LoRA variants tend to degrade performance---particularly on Healthcare, where LoRA-FullShot increases $\mathrm{MWQL}_n$ from $0.450$ to $0.593$.
 
\subsubsection{Summary of Key Findings}

Several patterns emerge from this analysis. First, \textbf{fine-tuning is generally beneficial}: with MWQL, zero-shot inference outperforms fine-tuning strategies in Chronos~2 on only five configurations out of 15, with only one corresponding to a long horizon; in TTM-R3-PT, it outperforms fine-tuning strategies on only four configurations out of 15, all of them on short or medium horizons. The situation is the same with MAPE for Chronos~2, but is more contrasted with MAPE for TTM-R3-PT, as zero-shot inference and fine-tuning strategies are preferred in eight and seven configurations out of 15, respectively. That said, as both models optimize a quantile loss close to MWQL, results evaluated with MAPE are more difficult to interpret; we believe that, on GIFT-Eval, TTM-R3-PT is more sensitive than Chronos~2 to the evaluation metric used. While generally beneficial, \textbf{fine-tuning is not universally beneficial}: on small datasets and certain domains (Nature, Energy), zero-shot inference can outperform all adaptation strategies, particularly for Chronos~2. Third, the \textbf{optimal adaptation strategy depends on model scale}: LoRA-FullShot is the most effective approach for Chronos~2, while FFT is better suited to TTM (a compact model). Fourth, \textbf{LoRA-FewShot is competitive in specific settings}---notably for Chronos~2 on Energy ($\mathrm{MAPE}_n$) and Transport (medium-term)---but generally underperforms LoRA-FullShot when sufficient data is available. Finally, \textbf{FFT is unreliable for Chronos~2}, consistently degrading performance across most settings, which underscores the risk of overfitting when fully updating a large pre-trained model on limited downstream data.

It is worth noting that, throughout our experiments, we used a single set of hyperparameters per model across all datasets (Table~\ref{tab:params}), without performing dataset-specific tuning. While per-dataset hyperparameter optimization would likely improve fine-tuning results in some cases, our fixed-budget protocol reflects a more realistic deployment scenario in which practitioners cannot afford extensive tuning for each new dataset. This choice may partly explain the underperformance of FFT on Chronos~2: a globally fixed learning rate and number of steps may be suboptimal for some datasets, exacerbating overfitting on smaller or less diverse ones. Conversely, it also highlights the robustness of LoRA-based adaptation, which proves effective under these constrained conditions without requiring careful per-dataset calibration.

%\subsubsection{Evaluation of Model Computational Efficiency}
%\textcolor{red}{TO BE UPDATED}
%We evaluate the efficiency of the Chronos, Moirai, and TimesFM models in terms of runtime and GPU memory consumption during inference, for different time series lengths (192 to 1984). The results, illustrated in Figure~\ref{fig:gp}, show that Moirai and TimesFM are faster than Chronos, while Moirai remains stable in GPU memory usage, unlike TimesFM and Chronos, whose consumption increases with series length.

%\begin{figure}[pos = t]
%    \centering
%    \includegraphics[width=13cm]{figs/image.png}
%    \caption{Efficiency analysis in terms of GPU memory (NVIDIA RTX A6000) and runtime during inference. The prediction horizon is 100 steps.}
%    \label{fig:gp}
%\end{figure}
%%%%%%%%%%%%%%%%%%%%%%%%%%%%%% Conclusion and Perspectives %%%%%%%%%%%%%%%%%%%%%%%%%%%%%%%%%%%%%%%%%%%%%%%%
\section{Conclusion and Perspectives}\label{sec:conc}
%\paragraph{Lack of high-quality data for pre-training.}

In this paper, we have presented a review of foundation models for time series forecasting, analyzing them through several lenses: architecture, pre-training, and adaptation via fine-tuning. We then investigated the effect of fine-tuning these models on target datasets, showing that, with an appropriate choice of hyperparameters, this step can significantly improve zero-shot performance—particularly for long-term forecasts, where zero-shot models tend to be less effective.

The field of foundation models for time series forecasting is still very recent and presents numerous challenges. On the one hand, unlike in NLP, time series data remain scarcely accessible and are often retained by large industrial companies, which limits the availability of vast and diverse datasets for research. On the other hand, foundation models applied to time series still lack theoretical and empirical studies that would enable a deeper understanding of their behavior in terms of generalization and information compression—especially given the strong heterogeneity of time series, which makes learning a universal forecaster particularly challenging.

%%%%%%%%%%%%%%%%%%%%%%%%%%%%%% appendix %%%%%%%%%%%%%%%%%%%%%%%%%%%%%%%%%%%%%%%%%%%%%%%%

\appendix

\printcredits
\section*{Acknowledgements}

The authors would like to thank Dr. Marwane Bouznif (Savoye) for insightful discussions and constructive feedback that helped improve this work.
\bibliographystyle{cas-model2-names}

\bibliography{cas-refs}


\end{document}